\documentclass[conference]{IEEEtran}
\usepackage{amsmath}
\usepackage{graphicx}
\usepackage{multirow}
\usepackage{multicol}
\usepackage{booktabs}
\usepackage{adjustbox}
\usepackage{xcolor}
\usepackage{xspace}
\usepackage{float}
\usepackage{subcaption}
\usepackage[hidelinks]{hyperref}

\title{Benchmarking fixed-length Fingerprint Representations across different Embedding Sizes and Sensor Types}
\author{\IEEEauthorblockN{T. Rohwedder$^2$, D. Osorio-Roig$^1$, C. Rathgeb$^1$, C. Busch$^1$}
\IEEEauthorblockA{1 - Biometrics and Security Research Group \\ Hochschule Darmstadt, Germany \\}
\IEEEauthorblockA{2 - Hochschule Darmstadt, Germany \\}
}

\def\eg{\textit{e.g.}\@\xspace} 
\def\ie{\textit{i.e.}\@\xspace}

\def\etal{\textit{et al.}\@\xspace} 

\begin{document}
\maketitle

\begin{abstract}
Traditional minutiae-based fingerprint representations consist of a variable-length set of minutiae. This necessitates a more complex comparison causing the drawback of high computational cost in one-to-many comparison. Recently, deep neural networks have been proposed to extract fixed-length embeddings from fingerprints. In this paper, we explore to what extent fingerprint texture information contained in such embeddings can be reduced in terms of dimension, while preserving high biometric performance. This is of particular interest, since it would allow to reduce the number of operations incurred at comparisons. We also study the impact in terms of recognition performance of the fingerprint textural information for two sensor types, \ie optical and capacitive. Furthermore, the impact of rotation and translation of fingerprint images on the extraction of fingerprint embeddings is analysed. Experimental results conducted on a publicly available database reveal an optimal embedding size of 512 feature elements for the texture-based embedding part of fixed-length fingerprint representations. In addition, differences in performance between sensor types can be perceived\footnote{As there is no public open-source code, the fingerprint texture representation algorithm will be made available after acceptance of the article.}.
\end{abstract}


\section{Introduction}
\label{sec:introduction}

Fingerprint recognition has been indispensable for decades in law enforcement and border control and the technology has been extended to numerous commercial applications. Recent market trends suggest that the popularity of fingerprint biometrics will increase further in the coming years~\cite{SkyQuest-FingerprintMarket-2022}, leading to broad deployment. This may further lead to higher workloads coupled with long transaction times.

The most commonly used fingerprint representation is based on minutiae. It is accurate and provides good interpretability of the ridge pattern of the fingerprint. Despite their popularity, minutiae-based representations lead to certain drawbacks, \eg variable length in terms of the number of minutiae and unordered feature vectors (\ie representation). A comparison of two minutiae sets commonly involves the determination of mated minutae pairs. This procedure can turn out to be computationally expensive, resulting in a complexity of $O(n^2)$~\cite[Chapter~4]{Maltoni-handbook-fingerprint-2022}. This computational complexity also limits the use of minutiae-based fingerprint representations in combination with biometric template protection (BTP) schemes (\eg homomorphic encryption~\cite{Bauspiess-FingerprintFHE-ICISSP-2023}) which results in a very high workload. Furthermore, their usability and interoperability are limited in feature-level multimodal fusion systems with other popular types of biometric characteristics (\eg face and iris) that use floating-point values in their representations (\eg~\cite{Boutros-elasticface-2022,Bauspiess-CoefficientPacking-FaceHE-IWBF-2022,Boutros-deep-iris-2022}).

\begin{figure*}[!t]
    \centering
    \includegraphics[width = \textwidth]{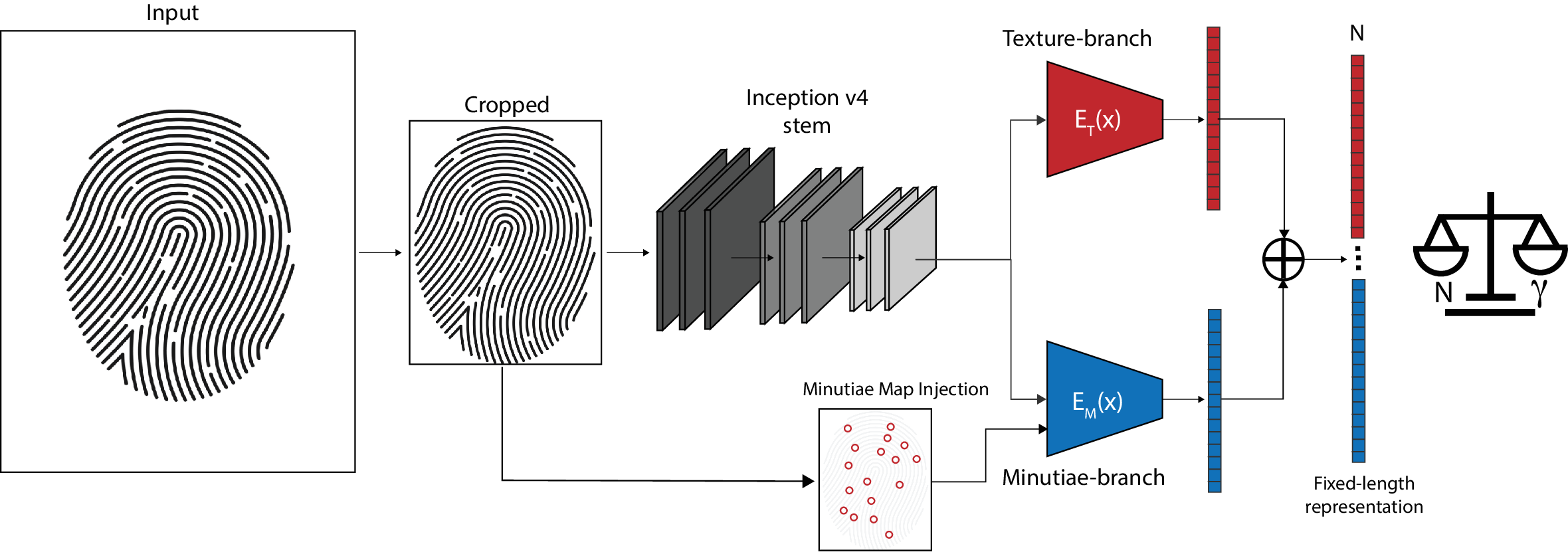}
    \caption{Conceptual overview of fixed-length fingerprint representation extraction. It consists of two branches; the upper branch $E_T(x)$ represents the fingerprint texture representation while the another one $E_M(x)$ is fed with the minutiae maps (\ie minutiae coordinates and angles) to learn a compact minutiae representation. The final fixed-length representation comprises the concatenation of $E_T(x)$ and $E_M(x)$.}
    \label{fig:overview}
\end{figure*}

In recent years, biometric technologies have been combined with deep learning approaches because of their capabilities to extract distinctive features, \ie embeddings, that allow high recognition performance~\cite{Liu-SphereFace-CVPR-2017,Wang-CosFace-CVPR-2018,Wang-deep-face-2021}. In particular, texture-based representation has been of interest for many types of biometric characteristics. Extracted texture information can easily be dimensionally reduced without sacrificing biometric performance (\eg search of the intrinsic dimensionality~\cite{Gong-intrinsic-2019} for face templates). In contrast to minutiae-based representations, the embeddings extracted by deep neural networks are usually of fixed length and, thus, can be successfully combined with BTP schemes and other types of biometric characteristics in a multimodal system. 

Recently, the extraction of texture-based fixed-length fingerprint representations has been proposed in different deep learning-based works~\cite{Engelsma-learning-2019,Grosz-transformerminutiae-2022}. Engelsma~\etal~\cite{Engelsma-learning-2019} proposed a Deep Neural Network (DNN) called DeepPrint that learns both minutiae and texture representations through multi-task learning. To evaluate the feasibility of the proposed approach, several experiments on some publicly available databases, \eg FVC 2004 DB1 A~\cite{Maio-FVC2004Competition-2004}, resulting in high recognition performance including scanned rolled fingerprints in NIST SD4~\cite{Watson-NISTSD4-1992} and NIST SD14~\cite{Watson-NISTSD14-1993} databases, were conducted by the authors. Despite the results achieved, a proper evaluation of this system based on different types of capture devices remains missing; only optical sensors are  assessed by the authors. In addition, there is still a lack of comprehensive research on the extent to which fingerprint texture representations can be dimensionally reduced without impairing recognition performance. Motivated by the above fact, this work explores the trade-off between dimensionality reduction and biometric performance for the competitive fixed-length representation extractor DeepPrint for data from optical and capacitive sensors.



The remainder of this paper is organised as follows: Sect.~\ref{sec:related_work} briefly introduces related works. In Sect.~\ref{sec:fixed-representation}, the considered method for extracting fixed-length fingerprint representation is explained in detail. Sect.~\ref{sec:experimental-setup} presents the experimental setup and the achieved results are summarised in Sect.~\ref{sec:results}. Final remarks are outlined in Sect.~\ref{sec:conclusions}.

\section{Related Work}
\label{sec:related_work}
The introduction of DNNs in biometrics in the last decade has led to the development of powerful face recognition systems which have replaced previously deployed schemes (\eg \cite{Deng-ArcFace-CVPR-2019}). Those architectures allow to derive fixed-length representations which contain the most significant facial traits representing the captured subject. In order to extend such scientific works, few articles have studied the feasibility of learning fixed-length fingerprint embeddings via DNNs~\cite{Engelsma-learning-2019,Grosz-transformerminutiae-2022}. Engelsma~\etal~\cite{Engelsma-learning-2019} proposed a scheme for learning texture-based fixed-length fingerprint representations. Using the domain knowledge injection of the minutiae map, the DNN approach produces a texture-based embedding of 192 components which reports competitive results in comparison with the one yielded by traditional minutiae-based techniques. Following this idea, Takahashi~\etal~\cite{Takahashi-CombinationMultitask-2020} included additional tasks on the multi-task learning framework proposed by Engelsma~\etal~\cite{Engelsma-learning-2019}. Subsequently, Grosz and Jain~\cite{Grosz-attention-maps-fingerprint-2022} introduced attention mechanisms within the DNN based on re-alignment strategies on local embeddings that refined the process of global embedding extraction. Afterwards, Grosz \etal~\cite{Grosz-transformerminutiae-2022} showed that the minutiae-based domain knowledge combined with vision transformers increased the biometric performance of the work in~\cite{Grosz-attention-maps-fingerprint-2022}.

\begin{figure}[!t]
    {\includegraphics[width = 0.5\textwidth]{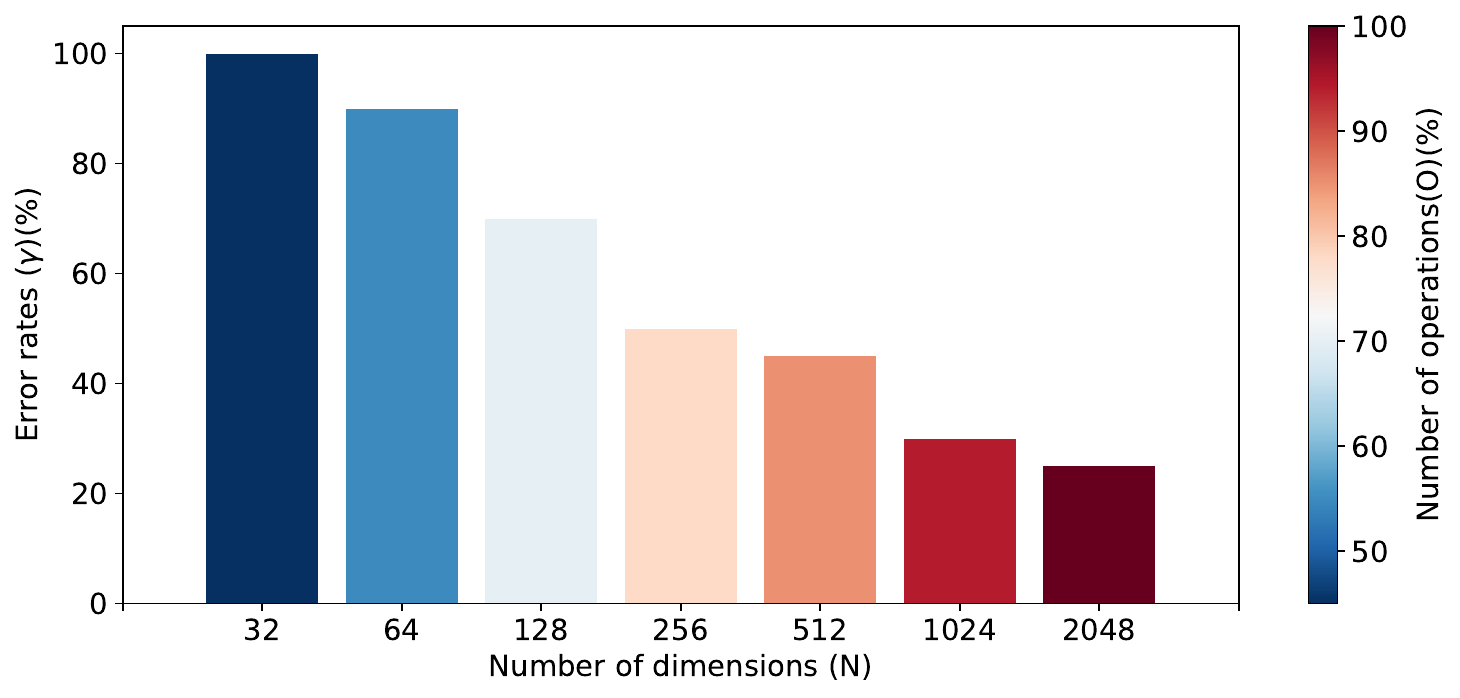}}
    {\caption{Relation of different values of $N$ (dimensions of embeddings) with respect to the biometric performance $\gamma$ and the number of operations $O$ performed at a single comparison. Consider that 100\% of $O$ represents the total of operations done by a fixed-length representation containing \eg $N$ = 2,048 floating points. In addition, $\gamma$ can represent some measure of evaluation in a biometric system (\eg FMR at 0.1\%).}
    \label{fig:balance_theoretical}}
\end{figure}

In general, previous approaches have focused on computing fixed-length discriminative representations from the fingerprint that achieve similar or superior biometric performance compared to traditional minutiae-based systems. Fixed-length representations can be easily combined with BTP schemes or used in a multi-modal pipeline and can therefore be deployed in  privacy-protecting authentication systems. This may result in significant differences in terms of texture appearance. Therefore, fixed-length representations must be robust to these sensor type variations. However, so far, the robustness towards different sensors has not yet been studied for fixed-length fingerprint representations.

\subsection{Deep fixed-length fingerprint representation}
\label{sec:fixed-representation}
Fig.~\ref{fig:overview} shows the overview of the DNN-based scheme used in this paper for learning the fixed-length fingerprint representation. As mentioned in Sect.~\ref{sec:introduction}, we selected and implemented the competitive DeepPrint approach in~\cite{Engelsma-learning-2019}. This system consists of two main branches. A cropped fingerprint image is initially processed by the stem of the Inception v4 architecture (henceforth referred to as ``stem''). Then the first branch $E_T(x)$, consisting of the remaining Inception v4 layers, performs the primary learning task of predicting a finger identity label. The second branch $E_M(x)$ also predicts the subject identity but it has a side task of detecting minutiae locations and orientations via the use of an AutoEncoder~\cite{Engelsma-learning-2019}. Thus, we guide this branch of the network to extract representations influenced by the fingerprint minutiae. The parameters of the stem are shared between the minutiae detection and representation learning branches. Finally, the embedding vectors computed by the two branches are concatenated to build the fingerprint representation. To train the network, we chose similar hyperparameters and loss functions as ~\cite{Engelsma-learning-2019}.

A drawback of our scheme is that the alignment step (prior to cropping the input data) was not considered. However, we investigated the effect of rotation and translation ( see Fig.~\ref{fig:heat-map-optical} and Fig.~\ref{fig:heat-map-capacitive}) of the fingerprint image on fixed-length representation for the two concatenated branches on different types of sensors. Also, as part of this work, we experimented with various embedding sizes, \ie $N = \{32, 64, 128, 256, 512, 1024, 2048\}$, and performed an ablation study of the complementary information provided by each branch. Theoretically, it is expected that values of $N$ have a significant impact on the biometric performance, while gaining workload (number of operations), as shown in Fig.~\ref{fig:balance_theoretical}.

\section{Experimental setup}
\label{sec:experimental-setup}

In this section, we describe the most relevant components in the training setup of the fixed-length extractor (Sect.~\ref{sec:training-db}) and databases used, while Sect.~\ref{sec:exp-metrics} describes metrics and protocols employed in the evaluation.


\begin{figure*}[!t]
\centering
\begin{subfigure}{0.20\linewidth}
    \includegraphics[width=\linewidth]{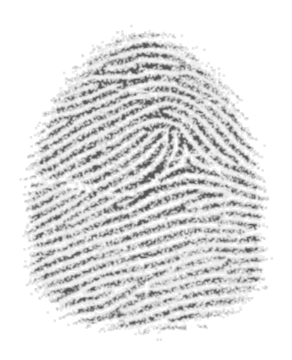}
    \caption{Original sample}
    \label{}
\end{subfigure}
\begin{subfigure}{0.20\linewidth}
    \includegraphics[width=\linewidth]{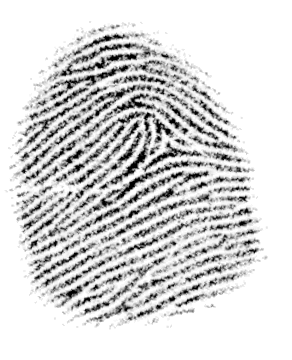}
    \caption{Data augmentation}
    \label{}
\end{subfigure}
\begin{subfigure}{0.20\linewidth}
    \includegraphics[width=\linewidth]{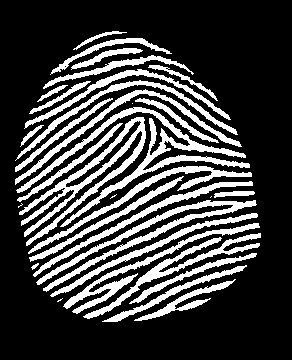}
    \caption{Enhanced image}
    \label{}
\end{subfigure}
\caption{Example of a synthetic fingerprint image generated by SFinGe with the respective pre-processing steps.}
\label{fig:training-preproc-sfinge}
\end{figure*}

\subsection{Training dataset}
\label{sec:training-db}

Although this work follows the architecture proposed by~\cite{Engelsma-learning-2019}, there are some other considerations such as the training database, and pre-processing steps that differ from the original work. In order to conduct the experimental analyses of this paper, a synthetic database for training the fixed-length approach was created. Note that the original database~\cite{Yoon-longitudinal-2015} used in~\cite{Engelsma-learning-2019} for training is not public. For the construction of the training database, synthetic fingerprint images together with the respective minutiae maps were generated by the framework SFinGe~\cite{Cappelli-sfinge-2004}. In particular, 40,000 samples stemming from 4,000 unique fingers were generated. To improve the generalisation capability of the network over real fingerprint images, 200 subjects from the MCYT database~\cite{Ortega-mcyt-2023}, which are equivalent to 200$\cdot$10 = 2,000 fingerprint instances, are selected and mixed into the set of synthetic fingerprint images forming a total of 40,000 + 48,000 samples. The variability of the different types of sensors together with the synthetic images generated is expected to contribute to a robust extraction of distinctive features. Note that, in our experiments, each fingerprint is considered a different biometric instance and is therefore assigned a different class in training time. The fingerprint images were cropped and resized to 299$\times$299 pixels as done in~\cite{Engelsma-learning-2019}. In addition, a data augmentation step based on the rotation, shifting, and variation of the brightness and contrast was randomly introduced. To enhance, the image quality, a pre-processing step was considered based on Gabor wavelet transformation~\cite{Karimimehr-gabor-2010}. Fig.~\ref{fig:training-preproc-sfinge} visualises examples of this virtual database composed with their corresponding pre-processing steps.

The remaining 130 different subjects from MCYT~\cite{Ortega-mcyt-2023} resulting 1,300 identities are used to evaluate the CNN-based approach. Tab.~\ref{tab:databases} summarises the characteristics of the training and testing database. 


\begin{table}[t!]
    \centering
    \caption{Summary of the training and testing sets employed in this work.}
    \begin{tabular}{c c c c c c} \toprule
      \multirow{2}{*}{\textbf{Databases}} &\multirow{2}{*}{\textbf{Sensors}}  & \multicolumn{2}{c}{\textbf{Training}}   &   \multicolumn{2}{c}{\textbf{Testing}}   \\ 
                                     &                 &   \#Subjects   &  \#Samples     &   \#Subjects   &  \#Samples      \\ \midrule
                        Synthetic    &    Optical      &   4,000        &   4,000$\times$10       &        -       &     -   \\  \midrule

    \multirow{2}{*}{MCYT~\cite{Ortega-mcyt-2023}}           &   Optical        &   200$\times$10  & 2,000$\times$12  &  130$\times$10   &  1,300$\times$12      \\
                                    &   Capacitive     &   200$\times$10  & 2,000$\times$12  &  130$\times$10   &  1,300$\times$12      \\

    \bottomrule
    \end{tabular}
    \label{tab:databases}
\end{table}

\subsection{Metrics and protocols}
\label{sec:exp-metrics}

Biometric performance in the verification scenario was reported in accordance with the metrics defined by ISO/IEC19795-1:2021~\cite{ISO-IEC-19795-1-Framework-210216}. The Equal Error Rate (EER), which represents the operating point at which False Match Rates (FMR) and False Non-Match Rates (FNMR) equalise, is computed. In addition, the FNMR values for several security thresholds, \ie 0 $\leq$ FMR $\leq$ 40 are depicted as Detection Error Trade-off (DET) curves. We also evaluate the identification rate for different rank values, \ie Rank-N on a closed-set identification scenario. Note that for the verification scenario, all possible comparisons for mated and non-mated comparisons are computed, while 10-fold cross-validation is performed on the closed-set identification protocol. Furthermore, the computational workload of a single comparison in relation to the embedding size (\ie feature dimensions ($N$)) and the number of operations ($O$) according to the cosine comparator are reported. Since the computed embeddings are normalised, the cosine similarity function between two fixed-length representations of size $N$ performs $N$ multiplications followed by $N - 1$ additions, resulting in $N \cdot (N - 1)$ operations.  

\begin{table}[t!]
	\scriptsize
	\begin{center}
	\caption{Biometric performance for the verification scenario using the texture-based branch. FNMR values are reported for a FMR at 0.1\%. The best result is highlighted in bold.}
	\label{tab:summary-embedding-dims}
\begin{adjustbox}{max width=\textwidth}
\begin{tabular}{cc  cc  cc }
\toprule
\multirow{2}{*}{\textbf{Embedding Size (N)}} & \multirow{2}{*}{\textbf{\#Operations (O)}}& \multicolumn{2}{c}{\textbf{Optical}} & \multicolumn{2}{c}{\textbf{Capacitive}} \\

           &              &            \textbf{FNMR} & \textbf{EER}          &   \textbf{FNMR} & \textbf{EER}  \\  \cmidrule{1-6}

32 & 63 & 5.36  & 1.19   & 10.60  & 2.42 \\
64 & 127 & 5.37 & 1.26    &  10.49  & 2.38  \\ 
128 & 255 & 3.51  & 1.00  &  7.91  & 2.14   \\ 
256 & 511 & 3.04  & 0.94   & 8.19  & 2.31   \\
\textbf{512} & \textbf{1023} & \textbf{1.89}  & \textbf{0.63}   &\textbf{5.39}& \textbf{1.68} \\ 
1024 & 2047 & 2.61 & 0.97  & 6.46  & 2.26    \\ 
2048 & 4095 & 3.91 & 1.17  &  9.46  & 2.51   \\ 
        
\bottomrule

\end{tabular} 	
\end{adjustbox}
    \end{center}
\end{table}

\section{Results}
\label{sec:results}

Tab.~\ref{tab:summary-embedding-dims} reports the biometric performance for different embedding sizes ($N$) as well as their respective number of operations ($O$). Following the theoretical behaviour presented in Fig.~\ref{fig:balance_theoretical}, we can observe that the biometric performance improves with $N$, resulting in the best performance $N$ = 512 (\ie EER = 0.63\% and EER = 1.68\% for optical and capacitive sensors, respectively). Note a slight degradation of performance for N $>$ 512, indicating the introduction of unreliable features at larger embeddings. Regarding the comparison between the performance depicted by both sensors, we perceive a significant deterioration in terms of EER and FNMR for the capacitive capture device. In particular, the algorithm for optical sensor images yields a FNMR@FMR = 0.1\% of 1.89\% for $N$ = 512, which is approximately three times lower than the one achieved on capacitive sensor images at the same security threshold (\ie FNMR = 5.39\%). These results demonstrate that the feature representation computed by the DeepPrint system is affected by the sensor technology of the fingerprint capture device. Therefore, further research to overcome this deficiency is necessary. 


\subsection{Performance analysis}

\begin{figure}[!t]
\centering
\begin{subfigure}{0.4\linewidth}
    \includegraphics[width=\linewidth]{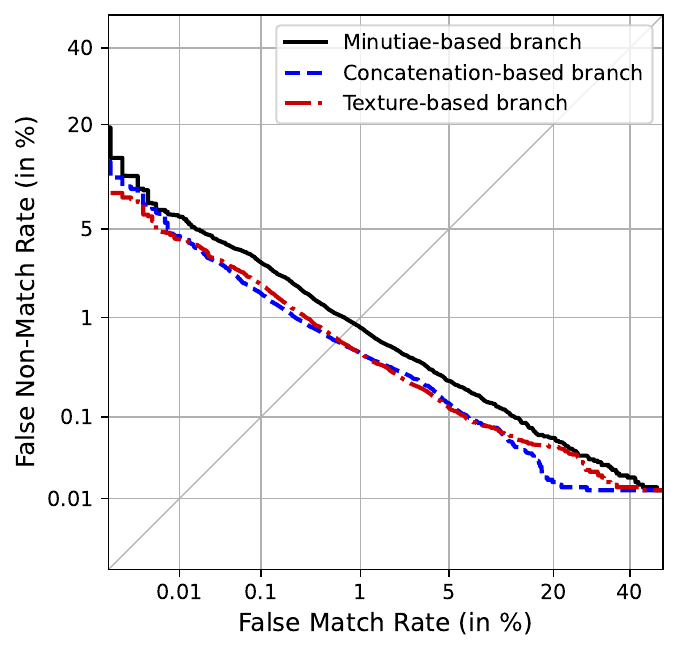}
    \caption{Optical sensor}
    \label{fig:optical_ver}
\end{subfigure}
\begin{subfigure}{0.4\linewidth}
    \includegraphics[width=\linewidth]{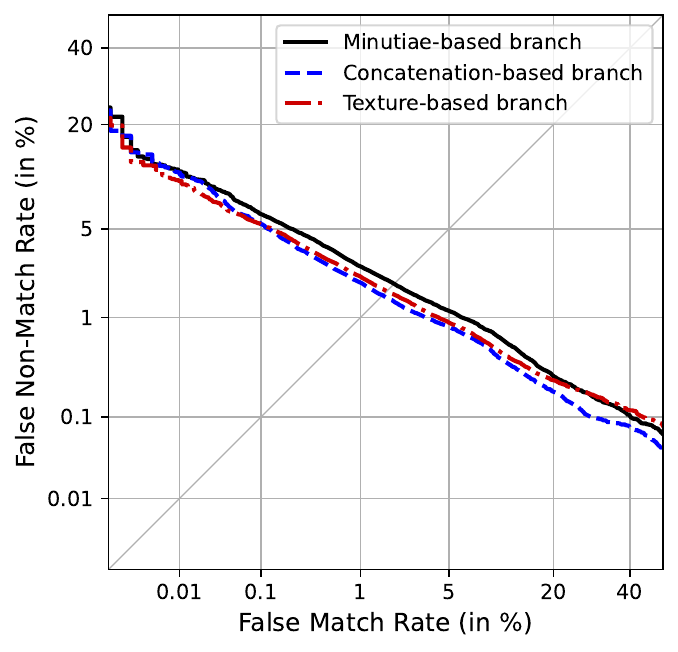}
    \caption{Capacitive sensor}
    \label{fig:capacitive_ver}
\end{subfigure}
\begin{subfigure}{0.45\linewidth}
    \includegraphics[width=\linewidth]{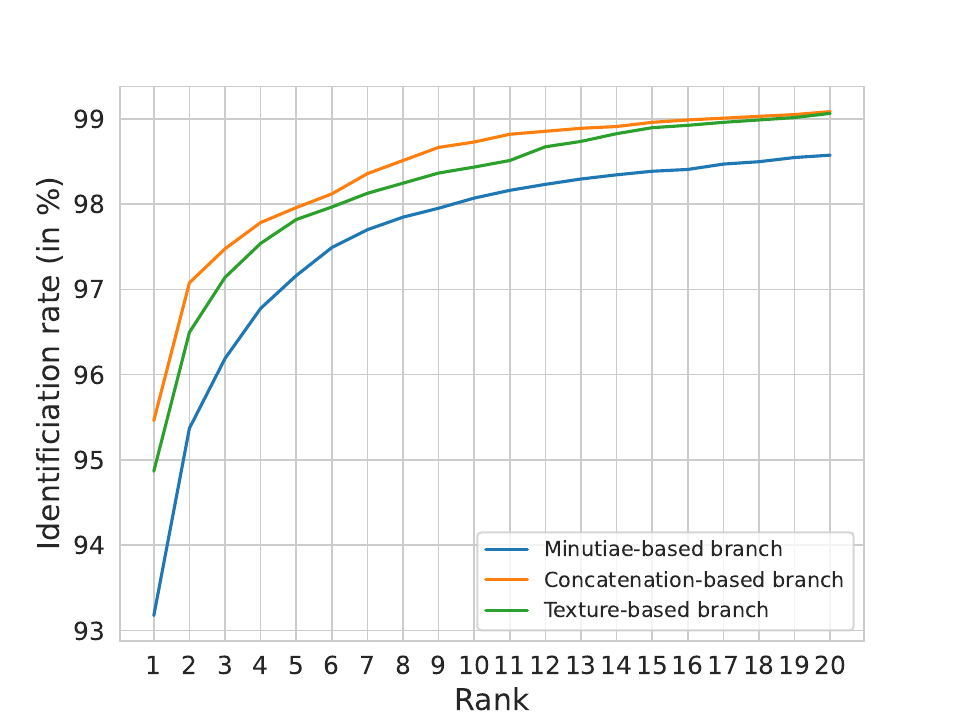}
    \caption{Optical sensor}
    \label{fig:optical_ident}
\end{subfigure}
\begin{subfigure}{0.45\linewidth}
    \includegraphics[width=\linewidth]{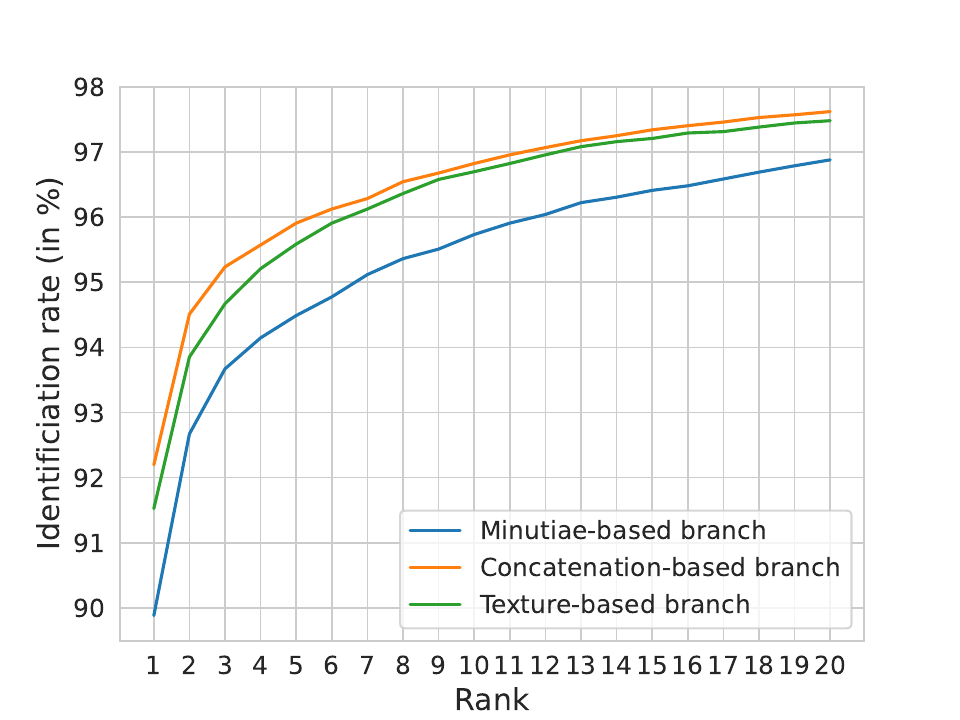}
    \caption{Capacitive sensor}
    \label{fig:capacitive_ident}
\end{subfigure}
\caption{DET curves for verification (a)-(b) and identification rate for different Rank values (c)-(d) on optical and capacitive sensors.}
\label{fig:res-variants}
\end{figure}

Fig.~\ref{fig:res-variants} depicts performance plots for $N$ = 512 for different branches, \ie minutiae, texture and concatenation-based branches, for both verification and closed-set identification scenarios. In this context, each evaluated branch is trained on a fixed-length embedding of 512 floating points. Note that both texture and the concatenation schemes in Fig.~\ref{fig:optical_ver} and Fig.~\ref{fig:capacitive_ver} achieve similar recognition performance, resulting in a similar FNMR below 2.0\% for optical sensor and FNMR = 5.0\% for the capacitive sensor at a FMR = 0.1\%. These non-significant differences over high-security thresholds (FMR$\leq$0.01\%) make the sole use of embeddings computed by the texture-based branch suitable to be combined with other approaches such as BTP and fusion schemes. It also avoids the detection of minutiae points which might lead to undesired recognition performance. On the other hand, we note, for the closed-set identification scenario in Fig.~\ref{fig:optical_ident} and Fig.~\ref{fig:capacitive_ident}, a slight improvement is obtained when the concatenation of minutiae and texture branches is performed. However, the differences between the identification rates reported by the concatenation and the texture-based branch is lower than 1.0\% at Rank-1 in the optical and capacitive sensors. We observe that the worst results are obtained with the minutiae-based branch, which results in a decrease of the identification rate down to 93\% and 90\% at Rank-1 for the optical and capacitive sensors, respectively. We believe that the non-considered alignment step in the optimisation of minutiae maps in the $E_T(x)$ branch leads to this performance deterioration. Despite this negative observation, we confirm that the concatenation of minutiae and texture-based embeddings complements each other to improve both independent branches. Finally, we note that similar to the results in Tab.~\ref{tab:summary-embedding-dims}, the evaluated system reports different recognition performances depending on the sensor employed.

\begin{figure}[!t]
\centering
\begin{subfigure}{0.45\linewidth}
    \includegraphics[width=\linewidth]{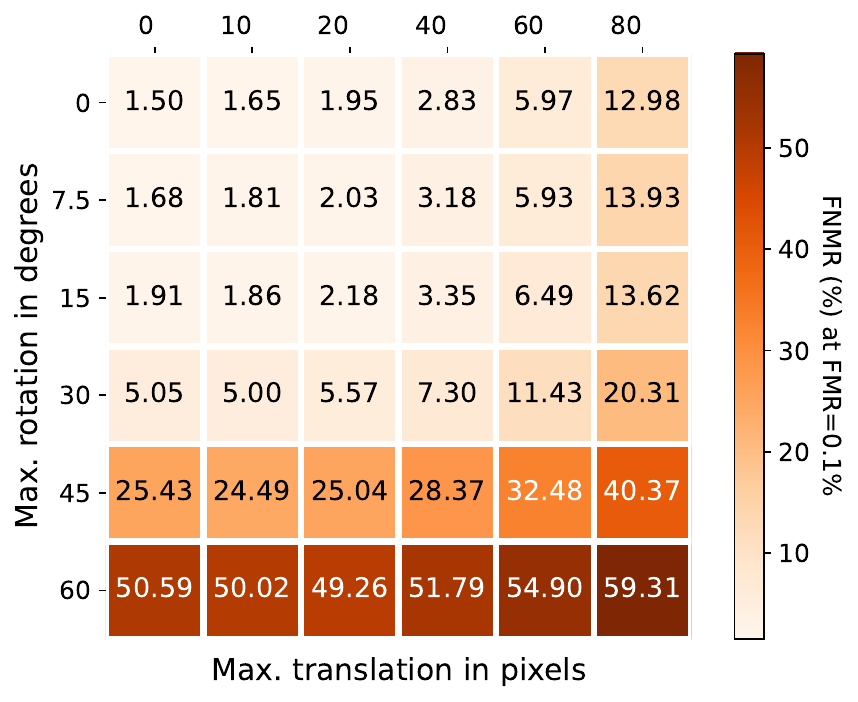}
    \caption{Optical sensor}
    \label{fig:heat-map-optical}
\end{subfigure}
\begin{subfigure}{0.45\linewidth}
    \includegraphics[width=\linewidth]{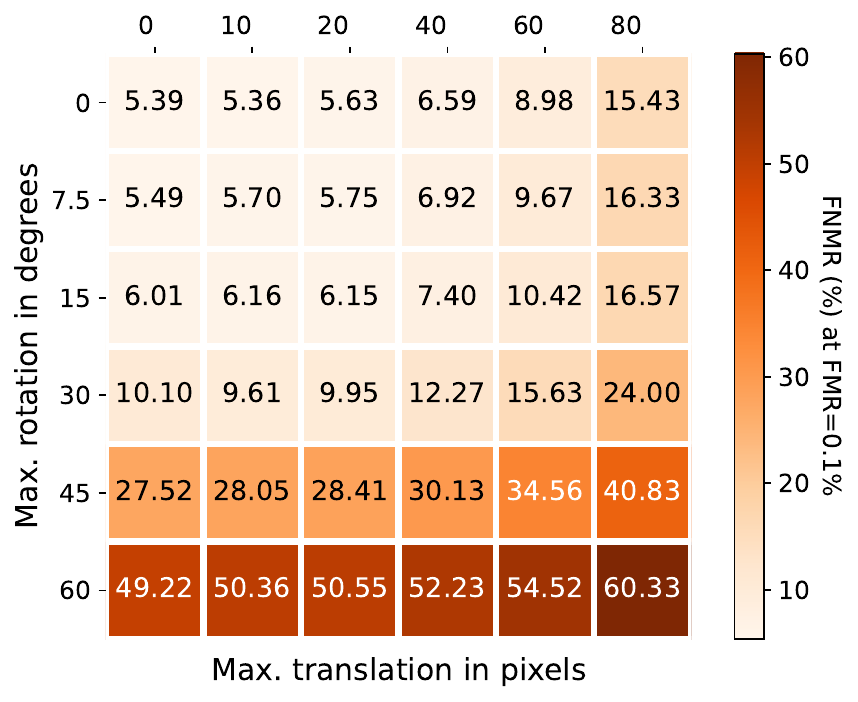}
    \caption{Capacitive sensor}
    \label{fig:heat-map-capacitive}
\end{subfigure}
\caption{Verification performance of the 512-dimensional concatenated embeddings for different levels of simulated rotation and translation. The value on the x-Axis is the maximum rotation $r$ where the input images were randomly rotated by a value sampled from the uniform distribution over $\protect[-r, r\protect]$. Respectively, on the y-Axis we have the max. translation $t$, where each fingerprint image is shifted by an amount of pixels sampled from $\protect[-t, t\protect]^2$.}
\label{fig:alignment-results}
\end{figure}

As mentioned in Sect.~\ref{sec:fixed-representation} the effect of the alignment of the fingerprint pose is explored in Fig.~\ref{fig:heat-map-optical} and Fig.~\ref{fig:heat-map-capacitive} for the optical and capacitive sensors, respectively. To that end, increasing levels of rotation and translation were applied to the testing set and then, fingerprint embeddings were extracted. Here, the performance (FNMR) deteriorates disproportionately as the magnitude of the rotation and translation increases.
Interestingly, the rate of deterioration appears to grow faster for rotation ($x\geq30$, $y=0$) compared to translation ($x=0$, $y\geq40$) for both sensors. These facts confirm the need for the alignment stage. Despite this, we believe that texture information contributes to some extent to reducing these negative effects. Future work should investigate this effect on the independent branches, including for minutiae map representations without texture information.

\section{Conclusions}
\label{sec:conclusions}
In this paper, we evaluated how dimensionality reduction for a state-of-the-art representation of fixed-length fingerprints affects the overall recognition performance. To do so, we analyse the degradation of biometric performance and computational workload in the comparison stage of the DeepPrint approach. Experimental results computing on a publicly available database empirically demonstrated that learned features with a dimension lower or higher than 512 floating points led to a  deterioration of biometric performance. Furthermore, the differences in performance between the results obtained for the optical and capacitive sensor indicated the need for further research in this field. In spite of this drawback, we do confirm that this fixed-length representation enables its use in combination with BTP schemes and multimodal schemes which is subject to our current research. 

\section{Acknowledgement}
This research work has been partially funded by the German Federal Ministry of Education and Research and the Hessian Ministry of Higher Education, Research, Science and the Arts within their joint support of the National Research Center for Applied Cybersecurity ATHENE, the European Union's Horizon 2020 research and innovation programme under the Marie Sk\l{}odowska-Curie grant agreement No. 860813 - TReSPAsS-ETN.  

\bibliographystyle{IEEEtran}
\bibliography{bibliography}

\end{document}